\documentclass[final,3p,times]{elsarticle}

\usepackage{lineno,hyperref}

\usepackage{mathptmx}
\usepackage{amssymb}
\usepackage{amsmath}
\usepackage{CJK}
\usepackage{algorithm}
\usepackage{algorithmicx}
\usepackage{algpseudocode}
\usepackage{booktabs}
\usepackage{multirow}

\usepackage{graphicx}
\usepackage{subfig}
\usepackage{makecell}
\usepackage{booktabs}

\begin{document}

\begin{frontmatter}

\title{Urban flows prediction from spatial-temporal data using machine learning: A survey}

\author[a,b,c]{Peng Xie}

\author[a,b,c]{Tianrui Li\corref{mycorrespondingauthor}}
\cortext[mycorrespondingauthor]{Corresponding author}
\ead{trli@swjtu.edu.cn}

\author[a,b,c]{Jia Liu}
\author[a,b,c]{Shengdong Du}
\author[a,b,c]{Xin Yang}
\author[d,a,b]{Junbo Zhang}

\address[a]{School of Information Science and Technology, Southwest Jiaotong University, Chengdu 611756, China}
\address[b]{Institute of Artificial Intelligence, Southwest Jiaotong University, Chengdu 611756, China}
\address[c]{National Engineering Laboratory of Integrated Transportation Big Data Application Technology, Southwest Jiaotong University, Chengdu 611756, China}
\address[d]{Urban Computing Business Unit, JD Finance, China}

\begin{abstract}
Urban spatial-temporal flows prediction is of great importance to traffic management, land use, public safety, etc. Urban flows are affected by several complex and dynamic factors, such as patterns of human activities, weather, events and holidays. Datasets evaluated the flows come from various sources in different domains, e.g. mobile phone data, taxi trajectories data, metro/bus swiping data, bike-sharing data and so on. To summarize these methodologies of urban flows prediction, in this paper, we first introduce four main factors affecting urban flows. Second, in order to further analysis urban flows, a preparation process of multi-sources spatial-temporal data related with urban flows is partitioned into three groups. Third, we choose the spatial-temporal dynamic data as a case study for the urban flows prediction task. Fourth, we analyze and compare some well-known and state-of-the-art flows prediction methods in detail, classifying them into five categories: statistics-based,  traditional machine learning-based, deep learning-based, reinforcement learning-based and transfer learning-based methods. Finally, we give open challenges of urban flows prediction and an outlook in the future of this field. This paper will facilitate researchers find suitable methods and open datasets for addressing urban spatial-temporal flows forecast problems.
\end{abstract}


\end{frontmatter}


\section{Introduction}

Forecasting urban flows is strategically important for traffic management, land use, public safety, etc. For city managers, they can pre-discover the traffic congestion that may occur in the city, deploy traffic in advance, and ease traffic congestion. For businessmen, they can find the crowded regions or potential business investment areas to gain greater business benefits. For public, they can improve their own travel plans in advance, stagger the peak of travel, and choose a more convenient way to travel. From the perspective of people's travel mode, urban flows contain crowd flow, traffic flow and public transit flow, etc.

However, urban flows prediction is not an easy issue. First, there are some main factors affecting urban flows, which can be classified into four groups. \emph{Daily flows activity patterns:} They are the main patterns of urban flows, including working day commute, going to and back from school and other daily repeated activities. \emph{Anomalies of flows activity patterns:} Although daily flows activity patterns are main patterns of urban flows, our mostly concern is anomalies of flows activity patterns and certain areas such as an increase of urban traffic anomaly, because this may lead to the phenomena of traffic congestion, social security, etc. For this kind of phenomenon, we can improve traffic deployment to make people travel more convenient and enhance security emergency to make social order more harmonious. Some urban events or activities will also affect the flows in the city. When temporary traffic control is imposed on an area because of the construction of roads, there will be a corresponding decrease in flows in that area. \emph{Weather:} It also has a certain impact on urban flows, such as heavy rain, smog and other extreme weather conditions. It causes the number of people going out may decrease, while the number of people going out may increase when the weather is sunny. \emph{Holidays:} In addition, for holidays such as National Day holiday and Spring Festival holiday, there may be cross-regional and surging flows, which is periodic in years. In terms of time, it also has a certain impact on the urban flows on the date near the holiday, making the fluctuation of crowd flows last for a period of time.

Datasets we can obtain are almost spatial-temporal data, such as mobile phone data, taxi trajectories data, metro/bus swiping data, etc., all of which have temporal dependence and spatial correlation. How to deal with spatial-temporal data is also a very valuable and fundamental problem.

In addition, predicting urban flows requires determining the level of forecasting, in terms of space, such as the entire city, regional, and street-wide level; in terms of time, such as the next 15 minutes, the next hour, the next 24 hours of urban flows in each regions. Different prediction levels require different precision and different processing methods. Apart from the objective factors mentioned above, there are also some unmeasured subjective travel intentions of the population, which are very difficult to study and haven't any breakthrough on it before.

The contribution of this paper lies in three folds. First, urban flows prediction from spatial-temporal data is systematically reviewed. Second, we divide the typical and representative methods into five categories for urban flows prediction and mainly analysis the deep learning-based methods. Third, some public spatial-temporal datasets for urban flows forecasting are shared for facilitating research.

The rest of this paper is organized as follows. In Section 2, we partition the multi-sources spatial-temporal data preparation process into three groups. In Section 3, we choose the spatial-temporal dynamic data as a case study for the urban flows prediction task. In Section 4, we analyze and compare some well-known and state-of-the-art flows prediction methods in detail, classifying them into five categories: statistics-based, traditional machine learning-based, deep learning-based, reinforcement learning-based and transfer learning-based methods. Finally, we give open challenges of urban flows prediction and an outlook on the future of this field.

\section{Data preparation}
Multi-sources data from urban must be processed and prepared for further data analysis. In this section, we partition the preparation process into three groups, they are shown as below.

\subsection{Spatial-temporal datasets}
The datasets used for urban flows prediction are most spatial-temporal data. From the characteristic perspectives of spatial and temporal, we can divide the datasets into three categories, which they are spatial-temporal static data, spatial static temporal dynamic data and spatial-temporal dynamic data. According to data type of the spatial-temporal datasets, there are point data and network data \cite{zheng2019urban}. The spatial-temporal datasets are illustrated in Table 1.

\begin{table}[h]
\caption{Types of spatial-temporal datasets.}
\centering
\normalsize
\resizebox{\textwidth}{10mm}{
\begin{tabular}{c|c|c|c}
\toprule[1pt]
	\textbf{Data type} &\textbf{Spatial-temporal static} &\textbf{Spatial static temporal dynamic} &\textbf{Spatial-temporal dynamic} \\
\midrule[1pt]
	\textbf{Point data} &POI and Events &Air quantity data and Road monitoring data&User check-in data and Phone signal data \\
	\textbf{Network data} &Road structure data &Traffic flow data and Crowd flow data &Taxi trajectory data and Public transit data\\
\bottomrule[1pt]
\end{tabular}}
\end{table}

\subsection{Map decomposition}
\begin{figure}
\centering
\includegraphics[height=10cm]{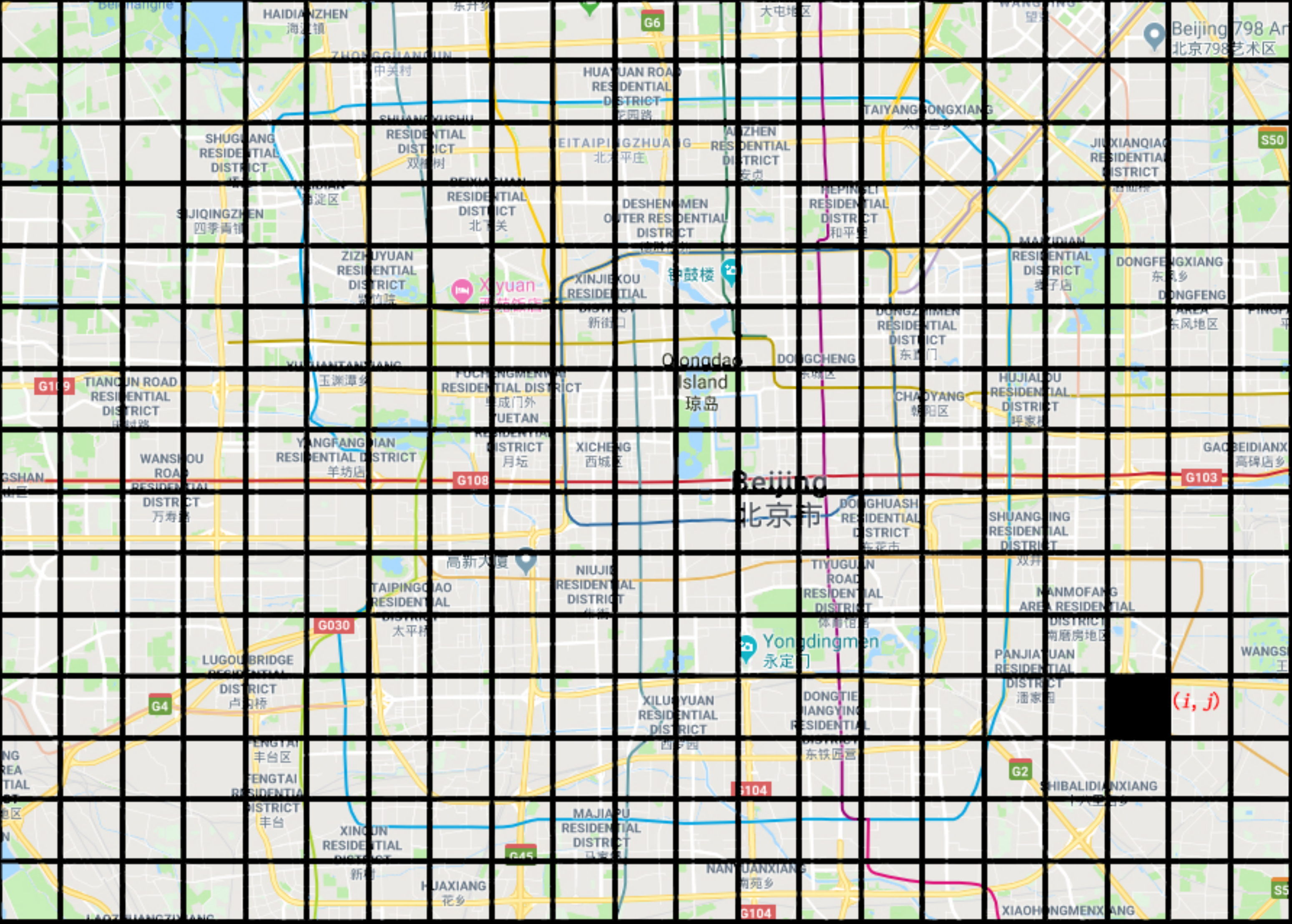}
\caption{\label{grid_map}Grid-based map segmentation in Beijing and the black region is an entertainment area (Happy Valley Beijing).}
\end{figure}

It can be found that in cities, a large amount of data is spatial-temporal data, such as traffic data including bicycle renting and returning data, taxi track data, metro card swiping data, etc. These data have time and space properties, and are in a constantly changing state. We need to represent and measure these data at the time and space level. In  order  to better process these spatial-temporal data, we should decompose the city map first. One  decomposition  method  is grid-based decomposition. For example, a DNN-based prediction model was proposed for spatial-temporal data \cite{zhang2016dnn}, which can capture both temporal and spatial properties. They defined a grid map based on the longitude and latitude, partitioned the Beijing city in to an \emph{M}*\emph{N} grip map. For example, there is an entertainment region (\emph{i}, \emph{j}) that lies at the \(i^{th}\) row and the \(j^{th}\) column in the grip map, as shown in Figure \ref{grid_map}. This is a good presentation of dividing region into some grids. Then they used in-flow and out-flow to measure the crowd flows in a region. Another is road network-based map decomposition method which the vehicles' GPS trajectories are mapped onto the city road network. Unlike the mentioned method before, it can sufficiently take advantage of the road network's information and apply classical clustering approach for further refining. Howerver, it is not as convenient and simple as grid-based decomposition method.

\subsection{Dealing with data problems}
When deal with spatial-temporal data, we may face many data problems. For example, data missing, data imbalance and data uncertainty arise alonely or in combination. These conditions will reduce the accuracy and efficiency of analysis result, so we review some methods to overcoming them.

\subsubsection{Data missing}
Due to sensor failures, communication errors, and other human factors, spatial-temporal data is often missing. Data missing often brings negative impact to the subsequent data analysis, so it is necessary to study the problem of data missing. At present, the main data missing processing method is to fill the missing value. For example, Lee et al. \cite{lee2008missing} proposed a factorial hidden Markov model to revover missing values. Hoang et al. \cite{hoang2016fccf} divided a city into low-level regions based on road network, and grouped adjacent these regions with similar crowd flow patterns using graph clustering. It's a novel and effective solution, but it's not sure how to define \emph{similar crowd flow patterns} accurately. To consider temporal and spatial correlation, Yi et al. \cite{yi2016st} proposed a spatial-temporal multi-view-based learning (ST-MVL) methods to collectively fill missing value in a collection of geo-sensory time series data. This method received good performance because of combining empirical statistic models, consisting of Inverse Distance Weighting and Simple Exponential Smoothing, with User-based and Item-based Collaborative Filtering. Hence, we can conclude that data missing in spatial-temporal datasets have an implicit spatiotemporal correlation.

\subsubsection{Data imbalance}
Spatiotemporal data imbalance is mainly manifested in two aspects: data distribution imbalance and data label imbalance. First of all, for the problem of imbalanced data distribution, Zheng et al. \cite{zheng2013u} proposed a semi-supervised learning algorithm to deal with the problem of sparse training data caused by the lack of air monitoring stations. Then, for the problem of imbalanced data label, Beckmann et al. \cite{beckmann2015knn} studied a KNN-based undersampling methods for data balancing. Wang et al. \cite{wang2018mutual} used a K-labelsets ensemble method based on mutual information and joint entropy to deal with inblanced data. Gong et al. \cite{gong2017rhsboost} presented a ensemble method using random undersampling and ROSE sampling to solve the imbalance classification problem. So when we face the data imbalance problem, it's a good choice to determine data distribution or data label imbalance, and then apply these corresponding methods.

\subsubsection{Data uncertainty}
In the actual deployment of machine learning algorithm, in order to better explain the model and effectively deal with the risk caused by data uncertainty, researchers proposed to adopt uncertainty quantification to alleviate the problem of data uncertainty \cite{khosravi2011comprehensive, begoli2019need}. Bayesian Deep Learning \cite{wang2016towards} is a kind of uncertainty quantification technique which can learn the weight distribution of networks. Quantifying predictive uncertainty in neural networks, which as a challenging problem, Lakshminarayanan et al. \cite{lakshminarayanan2017simple} proposed a model which puts uncertainty into the loss function and is directly optimized through BP algorithm. Rangapuram et al. \cite{rangapuram2018deep} combined state space models with deep learning for probabilistic time series forecasting, this method keeps properties of state space models such as data  interpretability. Recently, more and more uncertainty quantification research works emerge in the traffic forecast field, it could be hoped to be a novel direction for traffic forecast research.

\section{Spatial-temporal data for urban flows prediction}
In this section, we choose the spatial-temporal dynamic data (trajectories data) as a case study for the urban flows prediction task because it as one of widely studied spatial-temporal data types in urban flows forecasting and it relates more with our urban crowd flows and traffic flows prediction topic.

In the spatial-temporal data mining field, spatial-temporal (ST) data types can be divided into four categories \cite{atluri2018spatio}, which are different from the classification we mentioned in Table 1 above: (i) event data, which often occur at point locations and times (e.g., a concert or a car accident), (ii) trajectory data, which refer to the trajectories of moving objects (e.g., human, vehicle, animals), (iii) point reference data, where a continuous spatial-temporal field is being measured at moving ST reference sites (e.g., surface temperature are measured by using weather balloons), and (iv) raster data, whose measurements of an ST field are collected at fixed ST grids (e.g., air quality of Earth's surface collected by ground-based sensors). We can find that the ST events data and ST point reference data are point data, while trajectories data and ST raster data are network data. They can be merged into the categories of spatial-temporal datasets showed in Table 1. As we know, the derivation of trajectories data can be classified into four main categories, which are human mobility, transportation vehicles mobility, animals mobility and natural phenomena \cite{zheng2015trajectory}. In this paper, the urban flows prediction task mainly aim at estimated and predict human mobility and transportation vehicles mobility. Before starting data mining tasks for urban flows prediction, we should preprocess these spatial-temporary dynamic trajectory data. The raw location traces are often collected by smartphones with GPS and WiFi or taxis equipped with a GPS sensor. There are some spatial-temporal trajectory data preprocessing methods, consisting of extracting the history of place visits, data filtering and statistics, trajectory compression, trajectory segmentation and map matching \cite{zheng2015trajectory,do2015probabilistic}.

\begin{figure}
\centering
\includegraphics[height=4cm]{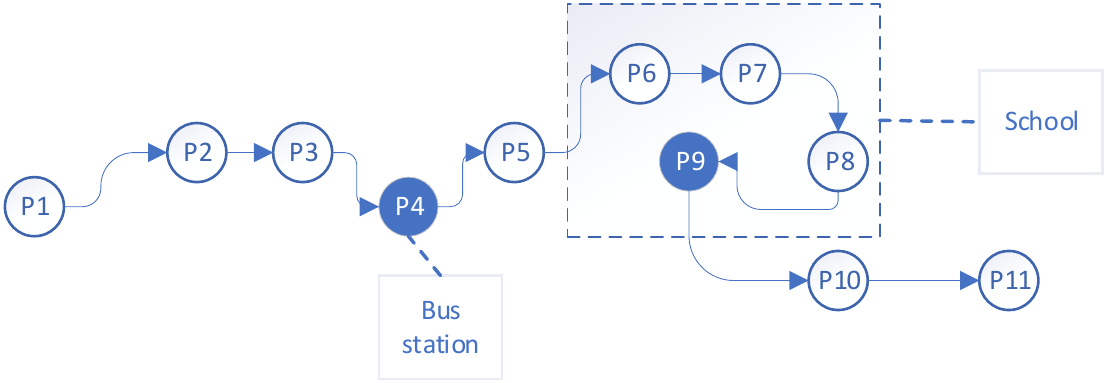}
\caption{\label{places_visited}Stay points and regions in trajectories.}
\end{figure}

\textbf{\emph{Extracting the history of place visits.}} From the analysis of the location traces, we can find that there are transitions and stay points in trajectories data. Then using a gird clustering algorithm \cite{do2015probabilistic}, the stay regions will be generated from the set of stay points based on a given radius. So the history of place that users visited can be extracted. As demonstrated in Figure \ref{places_visited}, if we set the minimum time of stay points to 5 min, the bus station (P4) can be found in this trajectory, and the stay region can be generated around the school (P9) about 1km radius. Based on these places with timestamp information, some applications may appear, such as travel recommendation, business location and travel time estimation.

\textbf{\emph{Data filtering and statistics.}} Because of sensor's error or other technical issues, there are some incomplete instances or outliers in the trajectories data. To forecast urban flows well, we need to filter out these incomplete instances and outliers before prediction task. The filter step is significant to avoid biased estimates of prediction performance. In order to find the trajectories distribution, we also need to perform some statistics analysis, such as mean, median, the ratio between location and transitions, and how many places visited by a user during a given recording interval.

\textbf{\emph{Trajectory compression.}} Let's imagine this condition. We can collect time-stamped location every second even more accurate time measurements for moving objects. It will cost plenty of communication, computing and storage sources. To efficiently collected and leverage these data, we need to compress the trajectory data. There are two major categories of trajectory compression methods. One is offline compression, such as Douglas-Peucker algorithm \cite{douglas2011algorithms} which replaces the original trajectory by an approximate line segment until the negligible error is below a specified error. The other is online compression, such as Sliding Window algorithm \cite{keogh2001online} and Open Window algorithm \cite{meratnia2004spatiotemporal} to transmit trajectory data timely. They are window-based algorithms which fit the trajectory points in a sliding window with a valid line segment and expand the sliding window until exceed specified error bound.

\begin{figure}
\centering
\includegraphics[height=1.5cm]{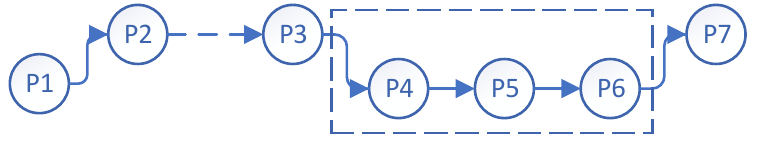}
\caption{\label{trajectory_segmentation}Time-interval and stay points trajectory segmentation.}
\end{figure}

\textbf{\emph{Trajectory segmentation.}} In order to classify or cluster trajectories to mine more useful knowledge, we need to study trajectory segmentation before mining tasks. There are three common types of trajectory segmentation methods. They are time interval-based, shape of a trajectory-based and semantic meaning-based methods. The first one is that a trajectory is divided into some segments based on a given time-length (lager than the given threshold or the same time interval), as illustrated in Figure \ref{trajectory_segmentation}. Due to the time length between p2 and p3 being larger than a given time interval, so we can divide the trajectory into two segments (p1p2 and p3p7). The second one is that we can partition a trajectory by the turning points with heading direction changing over a threshold \cite{zheng2015trajectory}. The last one is based on the semantic meaning of points in a trajectory. For example, in the travel speed estimation task, we often remove the stay points from the GPS trajectories because the stay points may be location where taxi is waiting for passengers \cite{yuan2012t}. For example, from Figure \ref{trajectory_segmentation}, the trajectory points in the dotted box can be removed because of stay points (p4p5p6) and the trajectory can be divided into two segments (p1p3 and p6p7). To find a walk-based segmentation \cite{zheng2010understanding,zheng2008learning}, we also need to combine with the human mobility patterns and employ further semantic meaning-based trajectory segmentation research.

\textbf{\emph{Map matching.}} There are two major categories map matching methods. One is the additional information-base method, and the other is the range of sampling points-based method. The first type of methods can be divided into four groups: geometric \cite{greenfeld2002matching}, topological \cite{chen2003integrated,yin2004weight}, probabilistic \cite{quddus2006high,pink2008statistical} and other advanced methods \cite{newson2009hidden,yuan2010interactive,lou2009map}. The second type of methods can be divided into two categories: local and global methods. The local methods aim to find a local optimal point based on the distance and orientation similarity. The global methods \cite{alt2003matching,brakatsoulas2005map} try to match an entire trajectory with a road network.

\section{Techniques for urban flows prediction}
Urban flows prediction is one of the spatial-temporal prediction tasks in urban computing field. There are some related research work, such as air quality prediction \cite{yi2018deep,li2016deep,li2017long}, traffic flow prediction \cite{du2018hybrid,polson2017deep,wu2018hybrid}, and travel demand prediction \cite{wang2018deepstcl,ke2017short,yao2018deep}. With the rapid development of Super first-tier city and the emergence of New first-tier city, urban flows prediction has become more and more important in traffic management and public safety. From the perspective of spatial forecasting measure, urban flows prediction can be divided into three categories, e.g., citywide-level, region-level and road-level \cite{hoang2016fccf,zhang2018predicting,jin2018spatio}. From the perspective of temporal prediction, urban flows prediction can be divide into three categories, e.g., short-term, mid-term and long-term flows prediction \cite{jin2018spatio,zhang2017deep}. And we can find that the problem of urban flows prediction has both spatial relation and temporal dependencies.

To solve the urban flows prediction problem, in recent years, there are many novel methods have been proposed. The major methods can be classified into five categories: statistics-based methods, traditional machine learning methods, deep learning-based methods, reinforcement learning methods and transfer learning methods.

\subsection{Statistics-based methods}
\begin{figure}
\centering
\includegraphics[height=12cm]{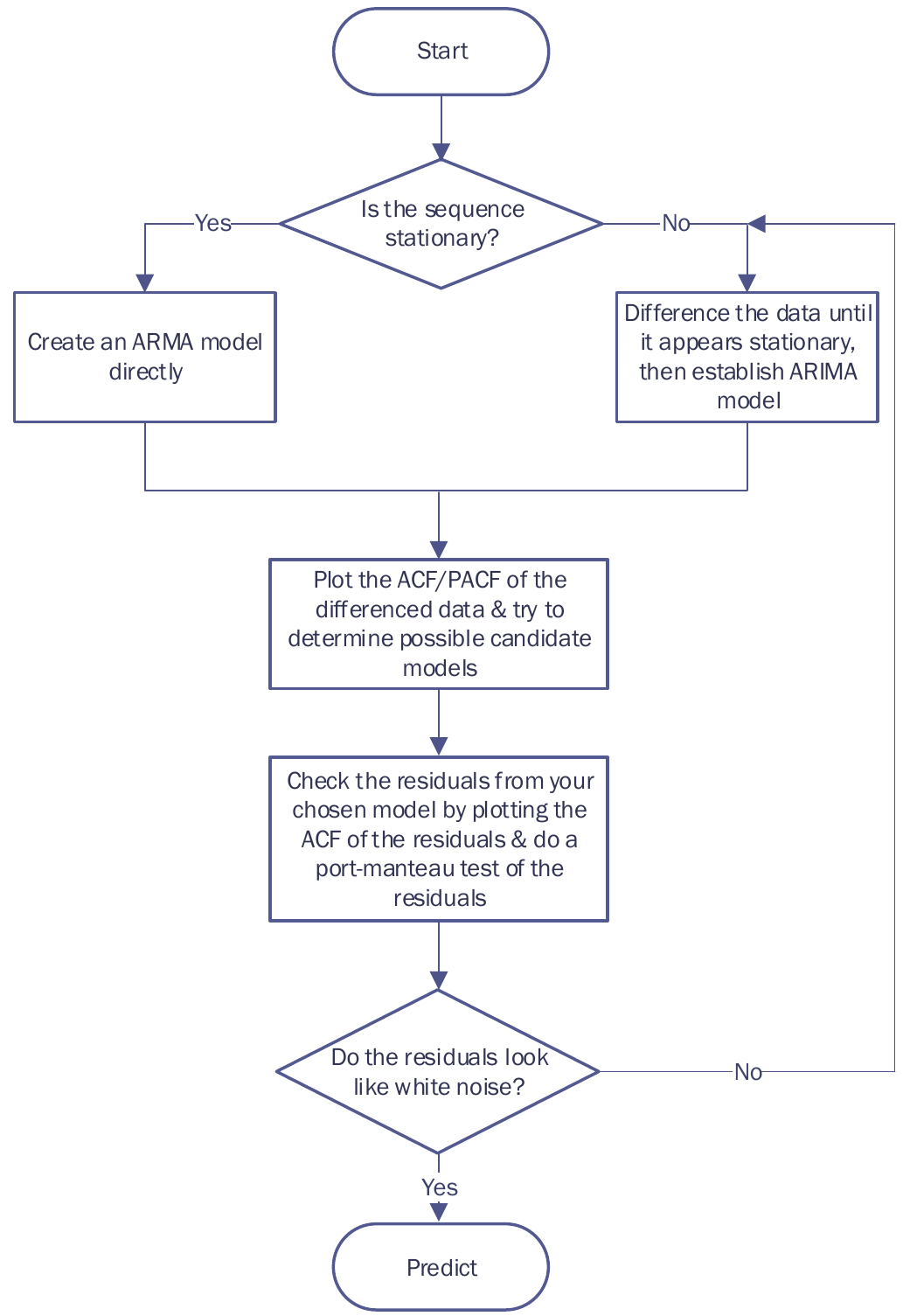}
\caption{\label{ARIMA}General process for forecasting using an ARIMA model.}
\end{figure}

In statistics-based methods, ARMA (Autoregressive Moving Average) \cite{said1984testing} is a fundamental time series prediction methods, and the variant method is ARIMA (Autoregressive Integrated Moving Average) \cite{williams1998urban}. An integrated version of ARMA model is also very popular in time-series prediction problems. You can use the Hyndman-Khandakar algorithm for automatic ARIMA modelling in R \cite{hyndman2007automatic}. The default procedure contains two steps \cite{hyndman2014forecasting} : (i) The number of differences $(0\leq d\leq 2)$ is determined using repeated KPSS tests; (ii) The value of \emph{p} and \emph{q} are then chosen by minimising the AICc (AICc is AIC (Akaike Information Criterion) with a correction for small sample sizes) after differencing the data \emph{d} times. Rather than considering every possible combination of \emph{p} and \emph{q}, the algorithm uses a stepwise search to traverse the model space. And the process for forecasting is summarized in Figure \ref{ARIMA}. As an another extension of the ARMA method, Seasonal Auto-Regressive Integrated Moving Average(SARIMA) method \cite{zhang2011seasonal} can catch intrinsic correlations in time series data, especially fit for modeling seasonal, stochastic time series that always occur in traffic flow data. Although these classical time-series methods can capture temporal dependencies in time series data, they can't depict the spatial influence in urban flows prediction problems.

\subsection{Traditional machine learning methods}
\begin{figure}
\centering
\includegraphics[height=5cm]{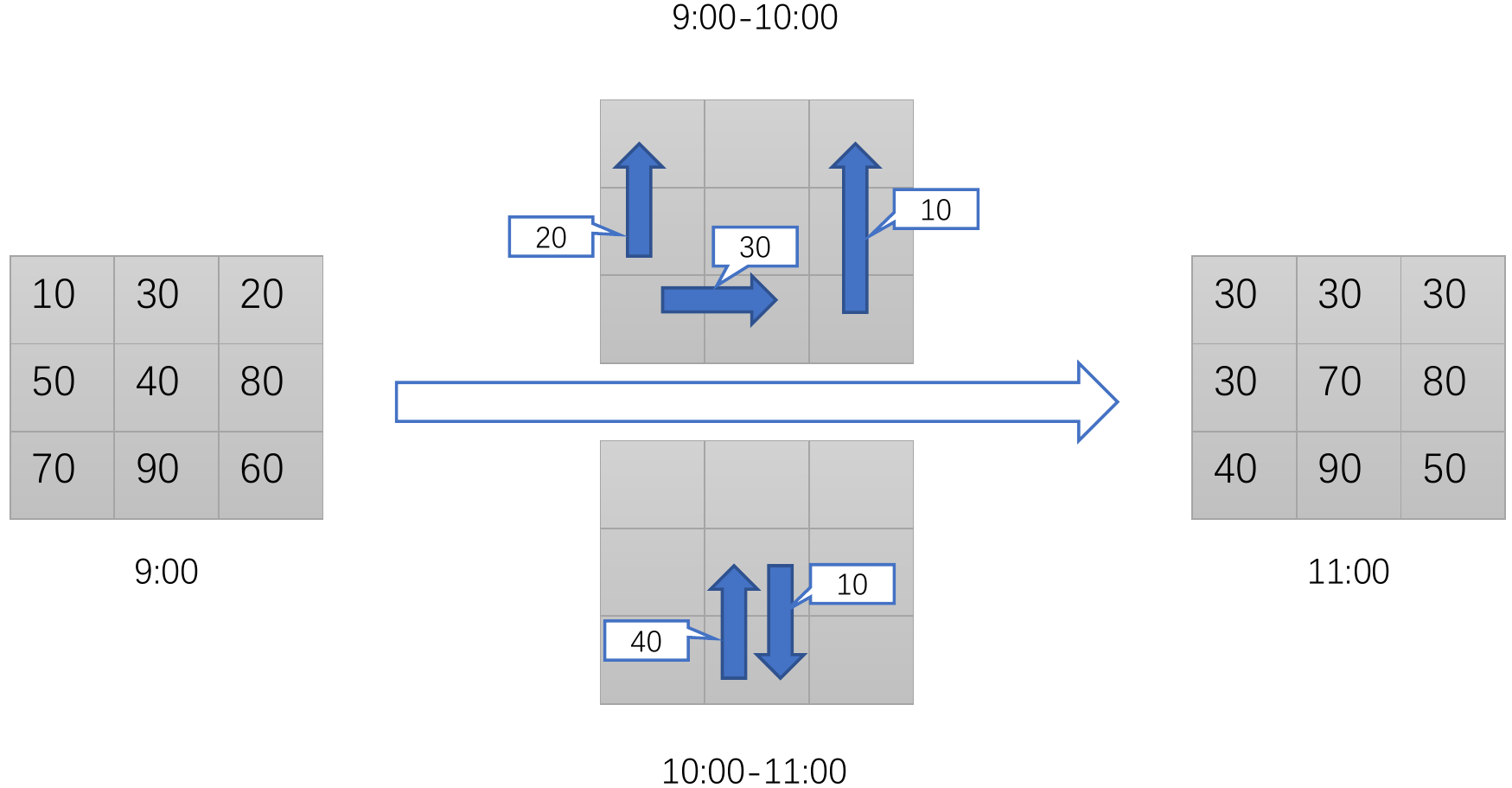}
\caption{\label{estimating people flow between cells}The task of estimating people flow between cells. Input: population of each grid cell at each time; Output: the number of people who move between cells over time. The map is divided by cells based on latitude and longitude.}
\end{figure}

Support vector regression (SVR) model is usually used in traffic flow prediction, for example, SVR with RBF kernel multiplied by a seasonal kernel has been used in traffic flow forecasting with high prediction accuracy and computational efficiency \cite{lippi2013short}. As one of non-parametric and data-driven methods, an enhanced K-nearest neighbor(K-NN) algorithm applied in short-term traffic flow prediction based on identify similar traffic patterns \cite{habtemichael2016short}. Zhu et al. studied a linear conditional Gaussian (LCG) Bayesian network(BN) model for short-term traffic flow prediction, which considers spatial-temporal characteristics as well as speed information \cite{zhu2016short}. To tackle the task of estimating the number of people who moved between cells, Akagi et al. \cite{akagi2018fast} developed a probabilistic model based on collective graphical models, which has considered movements to remote cells. As presented in Figure \ref{estimating people flow between cells}, the proposed method is an unsupervised learning method and only needs input variables, which are spatiotemporal population data. Liu et al. \cite{liu2018think} developed a graph processing framework based traffic estimation (GPTE), which can capture traffic correlation from taxi data and enable advanced traffic estimation at city-scale based on graph-parallel processing method. From these previous traditional machine learning-based methods, we can conclude that these models mainly focus on short-term traffic flow prediction and receive high prediction accuracy. However, traffic data explosion due to the increase of traffic sensors and the rapid development of intelligent transport systems in recent years, traditional machine learning-based methods are restricted with mining the deep, implicit spatial-temporal correlations in the big traffic data.

\subsection{Deep learning-based methods}
Deep learning-based methods are becoming main and popular methods for traffic spatial-temporal tasks. Due to big data and strong computing power, the success of deep learning in many application scenarios motivate plenty of deep learning-based methods in different areas, such as CNNs in computer vision \cite{lecun1998gradient,krizhevsky2012imagenet} and RNNs in sequence learning tasks \cite{williams1989learning,sutskever2014sequence}. In the next part, we will introduce the deep learning-based methods in detail.

In previous research work, there are many papers on human mobility prediction based on their history location trajectories data \cite{jiang2013review,calabrese2015urban,do2015probabilistic}. They aim at providing context-aware services and other location-based services for users, however, it may expose the privacy of users to others, so these data may be unavailable because of the policy of protecting privacy. Compared with the human mobility prediction problem, to forecast urban crowd flows, we can obtain more related datasets based on crowd-scale not individual-scale, e.g., taxi trajectories data, public transportation system data (metro or bus card swiping data), bike-sharing data, road network data, weather data and so on. And it is also of great importance to traffic management and public safety.

From the perspectives of people's travel mode and urban flows type, we can divide the urban flows into three categories: crowd flow, traffic flow, and public transit flow. Crowd flow usually can be concluded from users' phone signals and other vehicles GPS trajectories separately or synthetically. Traffic flow can mainly be estimated by using taxi trajectories data. Public transport flow indicates that the movement passengers measured by public transit card swiping data or bike-sharing data. If we want to measure the urban flows accurately, all the flow type need to be prepared and analyzed synthetically. However, in practical scenario, it's hard to receive all the urban flow type data of a city and there are some complex relationships among these flows. Next, we will review and analysis the three types of urban flows separately.

\subsubsection{Crowd flow prediction}
In recent years, there are many researchers focus on citywide-level traffic flow prediction \cite{hoang2016fccf,li2015traffic}, to forecast the citywide-level crowd flow. In the next part, before starting touch deep learning models, we will follow the related definitions of the crowd flow prediction problem in advance \cite{zhang2016dnn}.

\textbf{\emph{Region \cite{zhang2016dnn}.}} There are many definitions of Region in terms of different scales and semantic meanings. In most studies, they often partition a city into an I*J grid map based on longitude and latitude and a grid cell called a region, as shown in Figure \ref{grid_map}.

\textbf{\emph{Inflow/Outflow \cite{zhang2016dnn}.}} Let $\mathbb{P}$ be a collection of trajectories at the $t^{th}$ time interval. A grid cell (\emph{i}, \emph{j}) means the \(\mathit{i}^{th}\) row and the \(\mathit{j}^{th}\) column, the inflow and outflow of the crowds at the time interval \emph{ t } are defined respectively as

\begin{equation}
\label{Equ:inflow}
x_{t}^{in,i,j}=\sum_{T_{r}\in \mathbb{P}}\left | \left \{ k> 1|g_{k-1}\notin (i,j)\wedge g_{k}\in (i,j) \right \} \right |
\end{equation}

\begin{equation}
\label{Equ:outflow}
x_{t}^{out,i,j}=\sum_{T_{r}\in \mathbb{P}}\left | \left \{ k\geqslant 1|g_{k}\in (i,j)\wedge g_{k+1}\notin (i,j) \right \} \right |
\end{equation}
where \(T_{r}:g_{1}\rightarrow g_{2}\rightarrow ...\rightarrow g_{\left |T_{r} \right |}\) is a trajectory in \(\mathbb{P}\), and \(g_{k}\) is the geospatial coordinate, \(g_{k}\in (i, j)\) means the point \(g_{k}\) lies within grid (\emph{i}, \emph{j}) , and vice versa, \(\left | \cdot \right |\) denotes the cardinality of a set.

At the \emph{t}th time interval, inflow and outflow in all I*J regions denote as a tensor \(X_{t}\in \mathbb{R}^{2*I*J}\) where \(\left ( X_{t} \right )_{0,i,j}=x_{t}^{in,i,j}\), \(\left ( X_{t} \right )_{1,i,j}=x_{t}^{out,i,j}\).

\textbf{\emph{Prediction Target \cite{zhang2016dnn}.}} Given the historical observation \(\left \{ X_{t}|t=0,...,n-1 \right \}\), predict \(X_{n}\).

In 2016, a deep neural network-based prediction model called DeepST was proposed, which can capture spatial and temporal properties to predict citywide crowd flows \cite{zhang2016dnn}. The architecture of DeepST contains two parts: time sequence part and external factors part. From the history observation, the time serials contains temporal closeness, period and seasonal trend properties. The external factors have some related information with crowd flows prediction, e.g., dayofweek, weekday/weekend and meteorological condition. Then convolution layers are employed to capture the temporal closeness, period and seasonal trend properties, and the convolution layers output is fused followed by three sequential convolutional layers. At last, this result is fused with the output of the external factors captured by fully-connected layers and the prediction target \(X_{n}\) is obtained. And they built a real-time flow forecasting system (called as UrbanFlow) based on the DeepST model.

\begin{figure}
\centering
\includegraphics[height=10cm]{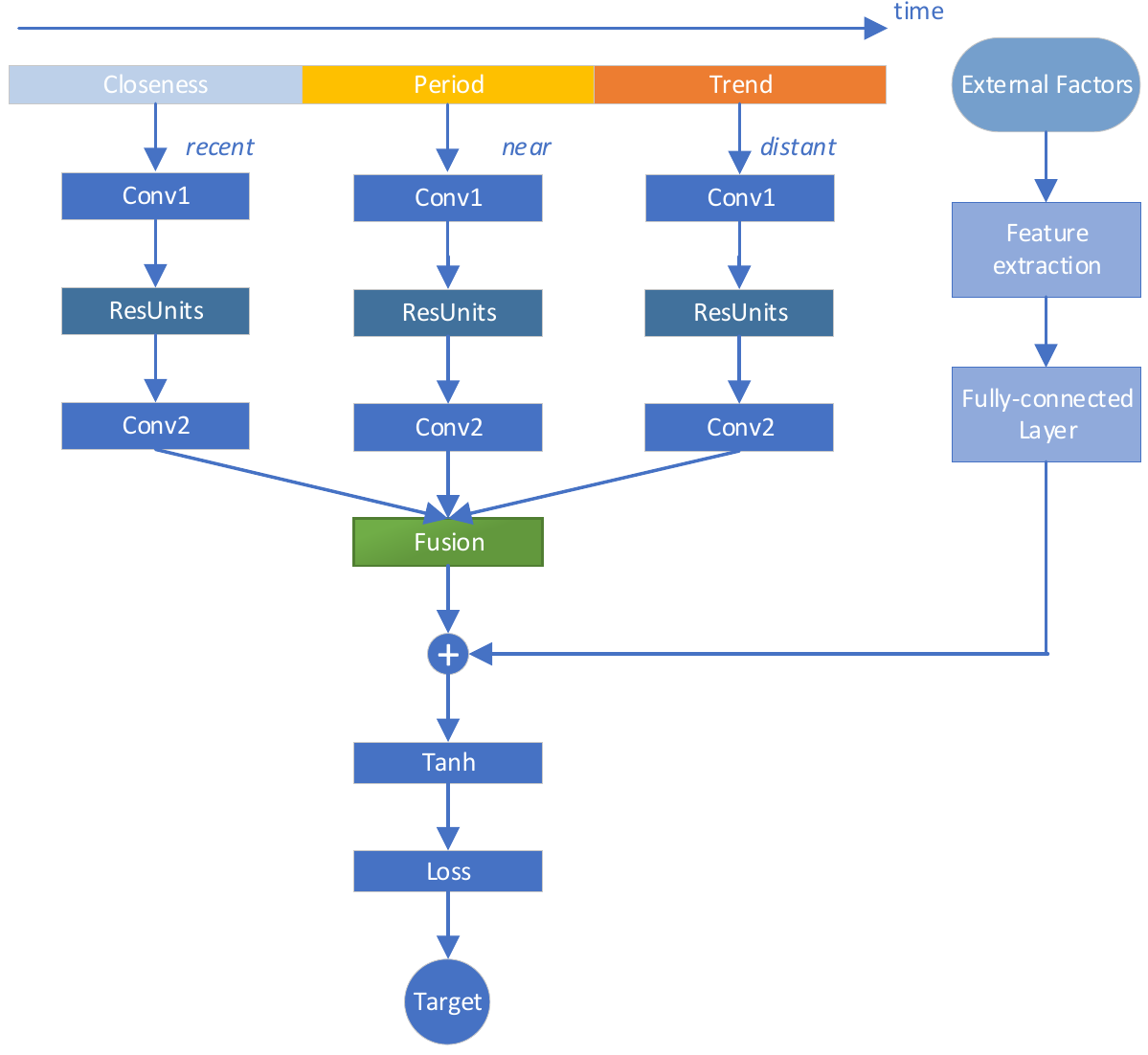}
\caption{\label{ST-ResNet}The architecture of ST-ResNet. (The original ST-ResNet architecture is available in \cite{zhang2018predicting}). The model outperforms other classical time-series and deep learning prediction methods.}
\end{figure}

\begin{figure}
\centering
\includegraphics[height=1.5cm]{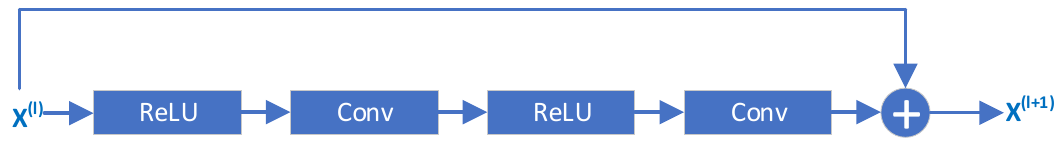}
\caption{\label{Res_Unit}Residual Unit. (The original Residual Unit is available in \cite{zhang2018predicting}).}
\end{figure}

In 2017 and 2018, a novel deep learning-based model (ST-ResNet) is presented shown in Figure \ref{ST-ResNet} and the city is partitioned by using a grid-based method for forecasting the crowd flows in each and every region of a city \cite{zhang2018predicting,zhang2017deep}. Note, The model outperforms other classical time-series and deep learning prediction methods. Mostly like the DeepST model, the ST-ResNet adds Residual Units and only one fusion component. The fusion step uses a parameter-matrix-based fusion method,

\begin{equation}
\label{Equ:fusion}
X_{Res}=W_{c}\circ X_{c}^{L+2}+W_{p}\circ X_{p}^{L+2}+W_{q}\circ X_{q}^{L+2}\in \mathbb{R}^{2*I*J}
\end{equation}
where \(\circ\) is Hadamard product (i.e., element-wise multiplication), \(W_{c}\), \(W_{p}\) and \(W_{q}\) are the learnable parameters that adjust the degrees affected by closeness, period and trend, respectively.
They concatenate the output of the three components after fusion with the external component. To model citywide dependencies, they employ residual learning in this ST-ResNet model, which has been demonstrated to be very effective for training super deep neural networks of over 1000 layers \cite{zhang2018predicting,he2016deep}. The residual unit used in the ST-ResNet is shown in Figure \ref{Res_Unit}. But in short-term crowd flows prediction problem, the residual network structure of ST-ResNet can be removed to get much more better performance, because it's not necessary to use the residual network structure to capture the distant spatial dependencies far away from the target region. And the ST-ResNet also needs too many data to train the model, so it has not good performance if we can't get much available data \cite{jin2018spatio}. Some researchers choose the ST-ResNet model as baseline to do further urban crowd flows prediction tasks \cite{zhang2018predicting,jin2018spatio}, because it outperforms other deep learning-based methods before.

\subsubsection{Traffic flow prediction}
Traffic flow prediction plays an important role and is a hot research field in urban flow forecasting \cite{deri2016big,zhan2017citywide,zhao2017lstm,duan2018improved}, because the taxi GPS data can access easily relatively and it can represent the citizens's traffic behaviors without the limitation of fixed lines. Fox example, Jiang et al. \cite{jiang2019geospatial} developed a deep learning framework which transforms geospatial data to images using Convolutional Neural Network(CNN) and residual networks for traffic prediction. Wu et al. \cite{wu2018hybrid} proposed a novel model, which combines the CNN and RNN to capture the spatial-temporal features as well as learn the importance of past traffic flow using attention mechanism. This method makes full use of the temporal and spatial characteristics of traffic flow to model and improve prediction performance. Since the forecast of traffic flow is affected by complex factors such as temporal relationship, spatial correlation, and other external factors (weather and events), it is more challenging to accurately predict traffic flow. Zhang et al. \cite{zhang2019flow} studied a multitask deep learning framework to simultaneously forecast the node flow and edge flow in the spatial-temporal networks. The model outperforms 11 baselines and shows great prediction performance in traffic flow forecast.

\subsubsection{Public transit flow prediction}
As an important component of the urban public transportation system, the metro has been rapidly deployed in the city because of its large capacity, high speed and high reliability, and it has attracted a large number of passengers \cite{liu2019deeppf,ma2018parallel,ning2018st,nam2017model}. Therefore, doing a good job of metro passenger flow forecast will not only help the metro management department to manage passenger travel demand and optimize metro dispatch, but also help passengers choose travel time and travel mode. Liu et al. \cite{liu2019deeppf} proposed an end-to-end deep learning model, named as DeepPF, to forecast the metro inbound and outbound passenger flow. They combine all the influence factors, such as temporal dependencies, spatial characteristics, metro operation properties and external environment factors to predict short-term metro passenger flow. The experiment shows that the model has good prediction performance and can apply to general conditions. We can find that it is feasible and effective to use the deep learning methods to capture the temporal and spatial characteristics of metro data and predict the metro passenger flow. Ma et al. \cite{ma2018parallel} analyzed the metro data's spatial-temporal characteristics and then developed a parallel framework which comprises convolutional neural network (CNN) and bi-directional long short-term memory network (BLSTM) to forecast metro passenger flow. The model was evaluated by Beijing metro network data and it outperformed traditional statistics methods. In recent years, bike-sharing is becoming more popular in urban transportation because of providing flexible transport mode and reducing the production of greenhouse gas. Chai et al. \cite{chai2018bike} proposed a novel multi-graph CNN method to predict bike flow at station-level. This method give us a novel graph neural network perspective to study traffic prediction.

\subsection{Reinforcement learning-based methods}
The reinforcement learning methods can usually be applied in traffic flow optimization problems. As we know, traffic congestion is a tricky problem in urban that may lead to travel delays, increased fuel consumption and air pollution. Hence, it is necessary to optimize traffic flow, make traffic control in advance, and alleviate traffic congestion. To tackle these challenges, Erwin et al. \cite{walraven2016traffic} proposed a method based on reinforcement learning, which to optimize traffic flow and using Q-learning to learn policies dictating the maximum driving speed allowed on a highway. The model takes traffic prediction into account and controls traffic flow proactively. More importantly, it can further help alleviate traffic congestion. Another example is to coordinate passenger inflow control problem on an urban rail transit line in Shanghai. In order to reduce the frequency of metro passengers being stranded and ensure public safety, Jiang et al. \cite{jiang2018reinforcement} presented a reinforcement learning-based method applied to study metro passenger inflow control strategy in peak hours. These methods aim at found better optimization strategy using reinforcement learning algorithms and further optimize current traffic flow, improve the efficiency of intelligent transportation systems. However, there are few articles that combine deep learning and reinforcement learning to predict and optimize traffic flow, it will be a promising research direction for the future.

\subsection{Transfer learning-based methods}
\begin{figure}
\centering
\includegraphics[height=7cm]{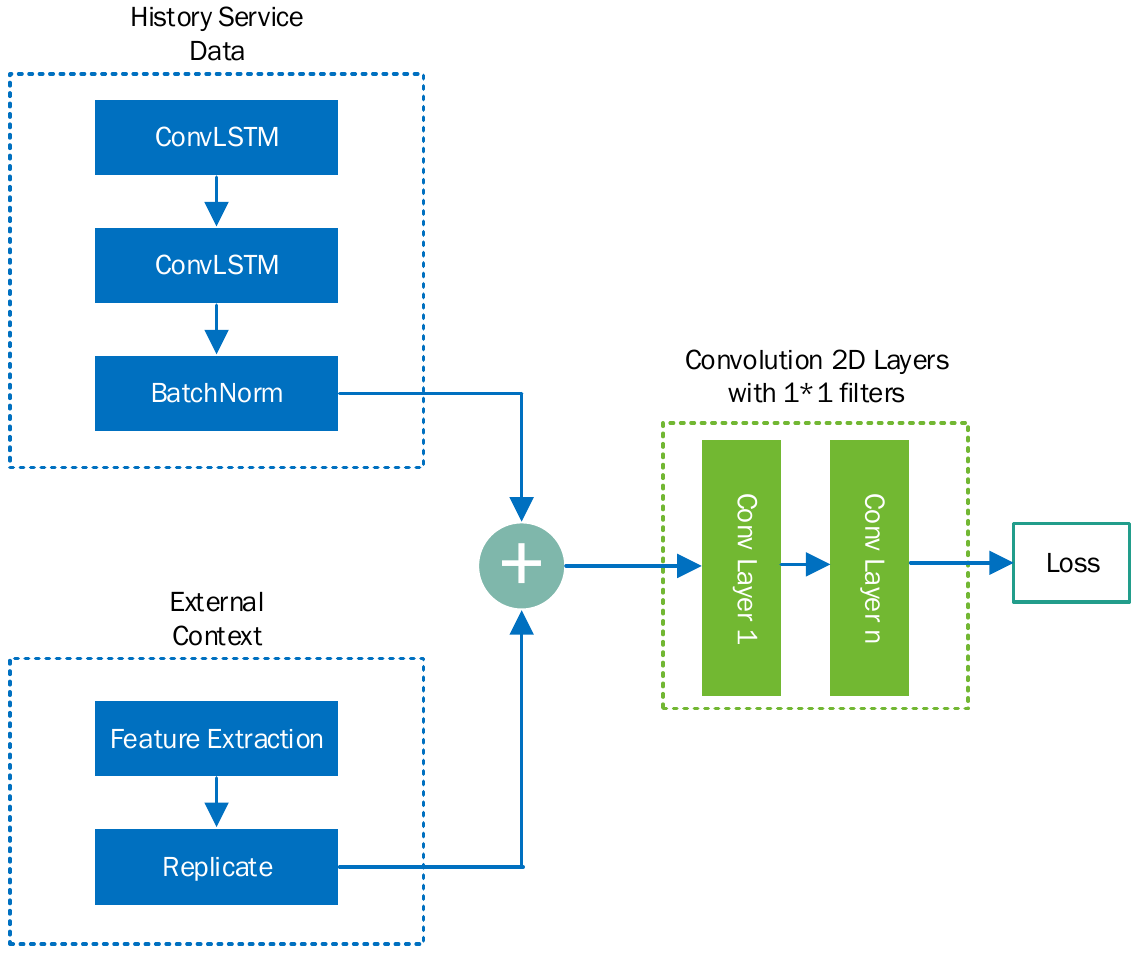}
\caption{\label{RegionTrans}Deep Spatial-temporal Neural Network with Region Representations. (The original model is available in \cite{wang2018crowd}).}
\end{figure}

Using transfer learning methods to predict crowd flows is the novel direction for urban flows forecast in a data-scarce city \cite{wang2018crowd,wang2018road}. A novel network for spatial-temporal prediction with region representation was developed \cite{wang2018crowd}, as shown in Figure \ref{RegionTrans}. The model's objective is to minimize the squared error between predicted \(\widetilde{y_{t}}\) and real \({y_{t}}\).

\begin{equation}
\label{Equ:min}
\min\limits_{\theta }\sum_{t\in \mathbb{T} }\left \| \widetilde{y_{t}}-y_{t} \right \|_{F}^{2}
\end{equation}

This method focuses on finding inter-city region pairs that share similar patterns and then transfers knowledge from data-adequate city (source city) to data-scarce city (target city). We can find that this method can use not too much data to predict the crowd flows than other deep learning-based methods and can take full advantage of knowledge from source city. But the model only shows good performance between two cities with similar patterns and it's not easy to find the matching function of the source regions and target regions.


\begin{table}[h]
\caption{A summary of previous works about urban flows prediction.}
\centering
\resizebox{\textwidth}{35mm}{
\normalsize
\begin{tabular}{c|c|c|c|c}
\toprule[1pt]
               \textbf{Urban flows type}    &\textbf{Task}  &\textbf{Dataset}   &\textbf{Method}   &\textbf{Reference}\\
\midrule[1pt]
\multirow{8}*{Crowd flow}
                   & Human mobility prediction &Mobile phone data	         &Probabilistic kernel method &\cite{do2015probabilistic} \\
                   & Forecasting citywide crowd flows	&Beijing taxi GPS data, NYC taxi trajectory data and NYC bike data 	&IGMRF and Bayesian network transit model	&\cite{hoang2016fccf}\\
                   & Crowd flow forecasting	&TaxiBJ15, TaxiGY16, LoopGY16 and BikeNYC14 	&DNN	&\cite{zhang2016dnn}\\
                   & Predicting citywide crowd flows 	&TaxiBJ and BikeNYC 	&CNN and Residual networks	&\cite{zhang2018predicting}\\
                   & Citywide short-term crowd flows prediction	&MobileBJ and TaxiBJ 	&CNN and LSTM	&\cite{jin2018spatio}\\
                   & Estimating People Flow	&GPS trace data of cars 	&A probabilistic model based on collective graphical models	&\cite{akagi2018fast}\\
                   & Crowd flow prediction	& Bike data of New York City and Chicago 	&Transfer learning-based framework	&\cite{wang2018crowd}\\
                   & Flow prediction in spatio-temporal networks	&TaxiBJ and TaxiNYC 	&Multitask deep-learning framework	&\cite{zhang2019flow}\\
\midrule[0pt]
\multirow{12}*{Traffic flow}
                   &Short-term traffic flow forecasting  &PeMS sensors data 	 &SVR  &\cite{lippi2013short}\\
                   & Short-term traffic flow prediction	&Traffic simulation data	&Linear conditional Gaussian Bayesian network	&\cite{zhu2016short}\\
                   & Taxi movement	&NYC taxi data and Road network data 	& Parallelization of the Dijkstra algorithm and statistics computations	&\cite{deri2016big}\\
                   & Traffic flow optimization	&Traffic simulation data	&Q-learning	&\cite{walraven2016traffic}\\
                   & Short-term traffic flow prediction	&Traffic flow data	&DNN	&\cite{polson2017deep}\\
                   & Short-term traffic forecast	&Traffic data	&LSTM	&\cite{zhao2017lstm}\\
                   & Estimate citywide traffic volume	&Road network data, POI data, GPS trajectories, Weather conditions and Videos clips	&A hybrid framework(machine learning techniques+traffic flow theory)	&\cite{zhan2017citywide}\\
                   & Short-term traffic flow forecasting	&UK traffic flow datasets	&A hybrid multimodal deep learning method	&\cite{du2018hybrid}\\
                   & Traffic flow prediction	& Traffic flow data	&CNN and GRU	&\cite{wu2018hybrid}\\
                   & Traffic flow prediction	&Taxi GPS trajectory data 	&CNN and LSTM  	&\cite{duan2018improved} \\
                   & Real-time traffic estimation at city-scale	&Road network data and SG taxi dataset 	&Graph-parallel processing framework	&\cite{liu2018think}\\
                   & Taxi pick-up/drop-off	&NYC taxi trip data 	&CNNs and Residual networks	&\cite{jiang2019geospatial}\\
\midrule[0pt]
\multirow{6}*{Public transit flow}
                   &Traffic prediction of  bike-sharing system  &Four datasets (bike data and meteorology data) from NYC and D.C 	 &Hierarchical prediction model  &\cite{li2015traffic}\\
                   &Metro stations crowd flows forecast   &Metro AFC record data  	 &Residual neural network framework  &\cite{ning2018st}\\
                   &Network-wide metro ridership prediction  &Metro ridership data  	 &CNN and BLSTM  &\cite{ma2018parallel}\\
                   &Optimize metro passenger inflow volume  &Metro data  	 &Reinforcement learning-based method  &\cite{jiang2018reinforcement}\\
                   &Bike flow prediction  &Bike data of New York city and Chicago, Weather data 	 &Multi-graph convolutional networks  &\cite{chai2018bike}\\
                   &Metro passenger flow prediction  &Metro data  	 &DNN  &\cite{liu2019deeppf}\\
\bottomrule[1pt]
\end{tabular}}
\end{table}

\section{Discussion}
With the development of machine learning algorithms, especially deep learning methods are still very active, and more open urban spatial-temporal datasets, many research works on urban flows forecast studies emerge during recent years. In order to give researchers a clear research progress outline in the field and help them with further research, we summarize the classic and representative works on urban flows forecast research during recent five years. We list these recent works mentioned in above Section 4, as shown in Table 2.

In Table 2, we categorize these works into three categories by tasks they faced, datasets and used methods. The abbreviations of some methods are given in Table 3. And then we will also list some open datasets to facilitate researchers further dig in the field.

From the Table 2, we can find that deep learning is still be frequently used in solving many urban flows forecast tasks, such as crowd flow prediction and traffic flow prediction. Due to more public transit data are becoming available on research under the privacy protection agreement, we can see that there are some public transit flow forecast papers using advanced machine learning algorithms. The works list in the Table 2 also give researchers good examples for dealing with urban spatial-temporal flow forecast problems. Learn from these great previous works, we can conclude that you can use statistics-based methods and traditional machine learning algorithms when addressing short-term traffic flow prediction tasks. If you want to capture urban flows' temporal dependency, spatial correlation simultaneously and further improve forecast performance, you can try to use deep learning-based methods, even a hybrid deep learning-based method, these methods have shown excellent results in spatial-temporal flow forecast tasks. Recently, reinforcement learning-based methods are often applied in urban traffic flow optimization problem, from few papers, we also see the method combines traffic flow prediction using deep learning and traffic flow optimization using reinforcement learning, it shows a promising direction for urban flows study. At last, if you have very little traffic data at hand, consider using a novel transfer learning approach. Compared to the data-hungry deep learning methods, it only requires a small amount of data to learn a lot because it has an ability to take advantage of the knowledge learned from the source domain.

\section{Open challenges}
In this article, we have reviewed how spatial-temporal data can be used to forecast urban flows. In dealing with these datasets and choosing great analysis methods, some challenges still remain challenges of urban spatial-temporal flows prediction are three-fold.

\textbf{\emph{Dealing with multiple influence factors.}} When we design our model, we may face following problems: what is the main factor affects urban spatial-temporal flows and what are the minor but necessary factors for our given tasks. For example, in this crowd flows prediction problem, we need consider the influence factors of spatial and temporal scale, and other factors, such as weather, holidays, social events, traffic accidents, traffic jams and so on. The main factors affecting urban flows discussed in the Introduction part of the paper may give you some inspirations.

\textbf{\emph{Finding suitable data fusion methods.}} In order to improve our model's performance, we often need more available datasets from various domains (traffic, weather, social) to train the deep learning-based model. However, we may face those problems: how to choose the suitable datasets for our tasks and how to fuse these heterogenous data as model input. Hence, it's not trivial to investigate cross-domain data fusion methods to speed up the research.

\textbf{\emph{Limitations in data sparsity.}} In practical scenario, some data are missing and failure due to sensor error, transmission failure and storage loss. These incomplete/inaccurate flow data will reduce the prediction accuracy. We can fill the missing value using some missing value filling methods. And transfer learning may be also a good choice to deal with the data sparsity challenge.

\begin{table}[h]
\caption{Abbreviations used in Table 2 and their corresponding terminologies (in order which they appear in Table 2).}
\centering
\begin{tabular}{c l}
\midrule[0.5pt]

               IGMRF    &Intrinsic gaussian markov random fields\\
               SVR     &Support vector regression\\
               DNN    &Deep neural network \\
               CNN    &Convolutional neural network\\
               LSTM    &Long short-term memory\\
               GRU    &Gated recurrent neural network\\
               BLSTM    &Bi-directional long short-term memory network\\
\bottomrule[0.5pt]
\end{tabular}
\end{table}

\section{Public urban spatial-temporal datasets}
In order to help other researchers further participate and make more valuable works, we collect and organize several related open datasets on this urban spatial-temporal forecast topic for you. Here are links of these datasets:
\begin{itemize}
  \item NYC taxi data: https://www1.nyc.gov/site/tlc/about/tlc-trip-record-data.page .
  \item NYC bike data: https://www.citibikenyc.com/system-data .
  \item San Francisco taxi data: https://crawdad.org/~crawdad/epfl/mobility/20090224/ .
  \item Weather and events data: https://www.wunderground.com/ .
  \item UK traffic flow datasets: http://data.gov.uk/dataset/highways-england-network-journey-time-and-traffic-flow-data .
  \item Traffic flow data is available from the Illinois Department of Transportation: http://www.travelmidwest.com/ .
  \item Weather and climate data: https://www.ncdc.noaa.gov/data-access .
  \item Chicago bike sharing data: https://www.divvybikes.com/system-data .
  \item NSW POI data: https://sdi.nsw.gov.au/catalog/search/resource/details.page?uuid=\%7BC41F6FE5-1C56-4556-9EC6-EC9BD7094BBB\%7D .
  \item Road network data: http://networkrepository.com/road.php .
\end{itemize}

\section{Conclusion and future work}
In this paper, we attempt to provide an overview of the field of urban spatial-temporal flows prediction during 2014-2019, which plays an increasingly significant role in urban computing research and has close relationship with traffic management, land use and public safety. However, we are only able to cover a small fraction of work in this rapid growing area of research. We still hope that this paper can provide you a ladder to do further research on urban spatial-temporal flows prediction. Due to most methods are data-driven methods in the urban flows prediction problem, we need to pay more attention to the data. This paper helps the reader in identifying problems with given spatial-temporal datasets and some good choice of preprocessing or prediction methods to deal with urban flows prediction problems. Although, the field of urban flows prediction has received much achievement, there are also many open challenges need to be dealt with, such as how to fuse multiple source data simultaneously, how to decide which influence factors are key factor for our problem and effectively solve the data sparsity problem. At last, we hope to see more generative and practical urban flows prediction methods in solving urban challenges.

\section*{Acknowledgments}
This research was supported by the National Natural Science Foundation of China (No. 61773324).


\bibliography{mybibfile}

\begin{thebibliography}{10}
\expandafter\ifx\csname url\endcsname\relax
  \def\url#1{\texttt{#1}}\fi
\expandafter\ifx\csname urlprefix\endcsname\relax\def\urlprefix{URL }\fi
\expandafter\ifx\csname href\endcsname\relax
  \def\href#1#2{#2} \def\path#1{#1}\fi

\bibitem{zheng2019urban}
Y.~Zheng, Urban Computing, MIT Press, 2019.

\bibitem{zhang2016dnn}
J.~Zhang, Y.~Zheng, D.~Qi, R.~Li, X.~Yi, Dnn-based prediction model for
  spatio-temporal data, in: Proceedings of the 24th ACM SIGSPATIAL
  International Conference on Advances in Geographic Information Systems, ACM,
  2016, p.~92.

\bibitem{lee2008missing}
D.~Lee, D.~Kulic, Y.~Nakamura, Missing motion data recovery using factorial
  hidden markov models, in: 2008 IEEE International Conference on Robotics and
  Automation, IEEE, 2008, pp. 1722--1728.

\bibitem{hoang2016fccf}
M.~X. Hoang, Y.~Zheng, A.~K. Singh, Fccf: Forecasting citywide crowd flows
  based on big data, in: Proceedings of the 24th ACM SIGSPATIAL International
  Conference on Advances in Geographic Information Systems, ACM, 2016, p.~6.

\bibitem{yi2016st}
X.~Yi, Y.~Zheng, J.~Zhang, T.~Li, St-mvl: Filling missing values in geo-sensory
  time series data, in: Proceedings of the Twenty-Fifth International Joint
  Conference on Artificial Intelligence, 2016, pp. 2704--2710.

\bibitem{zheng2013u}
Y.~Zheng, F.~Liu, H.-P. Hsieh, U-air: When urban air quality inference meets
  big data, in: Proceedings of the 19th ACM SIGKDD International Conference on
  Knowledge Discovery and Data Mining, ACM, 2013, pp. 1436--1444.

\bibitem{beckmann2015knn}
M.~Beckmann, N.~F. Ebecken, B.~S.~P. de~Lima, A knn undersampling approach for
  data balancing, Journal of Intelligent Learning Systems and Applications
  7~(04) (2015) 104.

\bibitem{wang2018mutual}
R.~Wang, S.~Kwong, Y.~Jia, Z.~Huang, L.~Wu, Mutual information based
  k-labelsets ensemble for multi-label classification, in: 2018 IEEE
  International Conference on Fuzzy Systems (FUZZ-IEEE), IEEE, 2018, pp. 1--7.

\bibitem{gong2017rhsboost}
J.~Gong, H.~Kim, Rhsboost: Improving classification performance in imbalance
  data, Computational Statistics \& Data Analysis 111 (2017) 1--13.

\bibitem{khosravi2011comprehensive}
A.~Khosravi, S.~Nahavandi, D.~Creighton, A.~F. Atiya, Comprehensive review of
  neural network-based prediction intervals and new advances, IEEE Transactions
  on Neural Networks 22~(9) (2011) 1341--1356.

\bibitem{begoli2019need}
E.~Begoli, T.~Bhattacharya, D.~Kusnezov, The need for uncertainty
  quantification in machine-assisted medical decision making, Nature Machine
  Intelligence 1~(1) (2019) 20.

\bibitem{wang2016towards}
H.~Wang, D.-Y. Yeung, Towards bayesian deep learning: A framework and some
  existing methods, IEEE Transactions on Knowledge and Data Engineering 28~(12)
  (2016) 3395--3408.

\bibitem{lakshminarayanan2017simple}
B.~Lakshminarayanan, A.~Pritzel, C.~Blundell, Simple and scalable predictive
  uncertainty estimation using deep ensembles, in: Advances in Neural
  Information Processing Systems, 2017, pp. 6402--6413.

\bibitem{rangapuram2018deep}
S.~S. Rangapuram, M.~W. Seeger, J.~Gasthaus, L.~Stella, Y.~Wang,
  T.~Januschowski, Deep state space models for time series forecasting, in:
  Advances in Neural Information Processing Systems, 2018, pp. 7796--7805.

\bibitem{atluri2018spatio}
G.~Atluri, A.~Karpatne, V.~Kumar, Spatio-temporal data mining: A survey of
  problems and methods, ACM Computing Surveys 51~(4) (2018) 1--37.

\bibitem{zheng2015trajectory}
Y.~Zheng, Trajectory data mining: An overview, ACM Transactions on Intelligent
  Systems and Technology (TIST) 6~(3) (2015) 1--41.

\bibitem{do2015probabilistic}
T.~M.~T. Do, O.~Dousse, M.~Miettinen, D.~Gatica-Perez, A probabilistic kernel
  method for human mobility prediction with smartphones, Pervasive and Mobile
  Computing 20 (2015) 13--28.

\bibitem{douglas2011algorithms}
D.~H. Douglas, T.~K. Peucker, Algorithms for the reduction of the number of
  points required to represent a digitized line or its caricature, Classics in
  Cartography: Reflections on Influential Articles from Cartographica (2011)
  15--28.

\bibitem{keogh2001online}
E.~Keogh, S.~Chu, D.~Hart, M.~Pazzani, An online algorithm for segmenting time
  series, in: Proceedings of the IEEE International Conference on Data Mining,
  IEEE, 2001, pp. 289--296.

\bibitem{meratnia2004spatiotemporal}
N.~Meratnia, A.~Rolf, Spatiotemporal compression techniques for moving point
  objects, in: Proceedings of the International Conference on Extending
  Database Technology, Springer, 2004, pp. 765--782.

\bibitem{yuan2012t}
J.~Yuan, Y.~Zheng, X.~Xie, G.~Sun, T-drive: Enhancing driving directions with
  taxi drivers' intelligence, IEEE Transactions on Knowledge and Data
  Engineering 25~(1) (2012) 220--232.

\bibitem{zheng2010understanding}
Y.~Zheng, Y.~Chen, Q.~Li, X.~Xie, W.-Y. Ma, Understanding transportation modes
  based on gps data for web applications, ACM Transactions on the Web 4~(1)
  (2010) 1--36.

\bibitem{zheng2008learning}
Y.~Zheng, L.~Liu, L.~Wang, X.~Xie, Learning transportation mode from raw gps
  data for geographic applications on the web, in: Proceedings of the 17th
  International Conference on World Wide Web, ACM, 2008, pp. 247--256.

\bibitem{greenfeld2002matching}
J.~S. Greenfeld, Matching gps observations to locations on a digital map, in:
  Proceedings of the 81th Annual Meeting of the Transportation Research Board,
  2002, pp. 164--173.

\bibitem{chen2003integrated}
W.~Chen, M.~Yu, Z.~Li, Y.~Chen, Integrated vehicle navigation system for urban
  applications, in: Proceedings of International Conference Global Navigation
  Satellite System, 2003, pp. 15--22.

\bibitem{yin2004weight}
H.~Yin, O.~Wolfson, A weight-based map matching method in moving objects
  databases, in: Proceedings of the International Conference on Scientific \&
  Statistical Database Management, IEEE, 2004, pp. 437--438.

\bibitem{quddus2006high}
M.~A. Quddus, R.~B. Noland, W.~Y. Ochieng, A high accuracy fuzzy logic based
  map matching algorithm for road transport, Journal of Intelligent
  Transportation Systems 10~(3) (2006) 103--115.

\bibitem{pink2008statistical}
O.~Pink, B.~Hummel, A statistical approach to map matching using road network
  geometry, topology and vehicular motion constraints, in: Proceedings of the
  11th International IEEE Conference on Intelligent Transportation, IEEE, 2008,
  pp. 862--867.

\bibitem{newson2009hidden}
P.~Newson, J.~Krumm, Hidden markov map matching through noise and sparseness,
  in: Proceedings of the 17th ACM SIGSPATIAL International Conference on
  Advances in Geographic Information Systems, ACM, 2009, pp. 336--343.

\bibitem{yuan2010interactive}
J.~Yuan, Y.~Zheng, C.~Zhang, X.~Xie, G.-Z. Sun, An interactive-voting based map
  matching algorithm, in: Proceedings of the 2010 Eleventh International
  Conference on Mobile Data Management, IEEE Computer Society, 2010, pp.
  43--52.

\bibitem{lou2009map}
Y.~Lou, C.~Zhang, Y.~Zheng, X.~Xie, W.~Wang, Y.~Huang, Map-matching for
  low-sampling-rate gps trajectories, in: Proceedings of the 17th ACM
  SIGSPATIAL International Conference on Advances In Geographic Information
  Systems, ACM, 2009, pp. 352--361.

\bibitem{alt2003matching}
H.~Alt, A.~Efrat, G.~Rote, C.~Wenk, Matching planar maps, Journal of Algorithms
  49~(2) (2003) 262--283.

\bibitem{brakatsoulas2005map}
S.~Brakatsoulas, D.~Pfoser, R.~Salas, C.~Wenk, On map-matching vehicle tracking
  data, in: Proceedings of the 31st International Conference on Very Large Data
  Bases, VLDB Endowment, 2005, pp. 853--864.

\bibitem{yi2018deep}
X.~Yi, J.~Zhang, Z.~Wang, T.~Li, Y.~Zheng, Deep distributed fusion network for
  air quality prediction, in: Proceedings of the 24th ACM SIGKDD International
  Conference on Knowledge Discovery and Data Mining, 2018, pp. 965--973.

\bibitem{li2016deep}
X.~Li, L.~Peng, Y.~Hu, J.~Shao, T.~Chi, Deep learning architecture for air
  quality predictions, Environmental Science and Pollution Research 23~(22)
  (2016) 22408--22417.

\bibitem{li2017long}
X.~Li, L.~Peng, X.~Yao, S.~Cui, Y.~Hu, C.~You, T.~Chi, Long short-term memory
  neural network for air pollutant concentration predictions: Method
  development and evaluation, Environmental Pollution 231 (2017) 997--1004.

\bibitem{du2018hybrid}
S.~Du, T.~Li, X.~Gong, Z.~Yu, Y.~Huang, S.-J. Horng, A hybrid method for
  traffic flow forecasting using multimodal deep learning, arXiv preprint
  arXiv:1803.02099.

\bibitem{polson2017deep}
N.~G. Polson, V.~O. Sokolov, Deep learning for short-term traffic flow
  prediction, Transportation Research Part C: Emerging Technologies 79 (2017)
  1--17.

\bibitem{wu2018hybrid}
Y.~Wu, H.~Tan, L.~Qin, B.~Ran, Z.~Jiang, A hybrid deep learning based traffic
  flow prediction method and its understanding, Transportation Research Part C:
  Emerging Technologies 90 (2018) 166--180.

\bibitem{wang2018deepstcl}
D.~Wang, Y.~Yang, S.~Ning, Deepstcl: A deep spatio-temporal convlstm for travel
  demand prediction, in: 2018 International Joint Conference on Neural
  Networks, IEEE, 2018, pp. 1--8.

\bibitem{ke2017short}
J.~Ke, H.~Zheng, H.~Yang, X.~M. Chen, Short-term forecasting of passenger
  demand under on-demand ride services: A spatio-temporal deep learning
  approach, Transportation Research Part C: Emerging Technologies 85 (2017)
  591--608.

\bibitem{yao2018deep}
H.~Yao, F.~Wu, J.~Ke, X.~Tang, Y.~Jia, S.~Lu, P.~Gong, J.~Ye, Deep multi-view
  spatial-temporal network for taxi demand prediction, arXiv preprint
  arXiv:1802.08714.

\bibitem{zhang2018predicting}
J.~Zhang, Y.~Zheng, D.~Qi, R.~Li, X.~Yi, T.~Li, Predicting citywide crowd flows
  using deep spatio-temporal residual networks, Artificial Intelligence 259
  (2018) 147--166.

\bibitem{jin2018spatio}
W.~Jin, Y.~Lin, Z.~Wu, H.~Wan, Spatio-temporal recurrent convolutional networks
  for citywide short-term crowd flows prediction, in: Proceedings of the 2nd
  International Conference on Compute and Data Analysis, ACM, 2018, pp. 28--35.

\bibitem{zhang2017deep}
J.~Zhang, Y.~Zheng, D.~Qi, Deep spatio-temporal residual networks for citywide
  crowd flows prediction, in: Proceedings of the Thirty-First AAAI Conference
  on Artificial Intelligence, 2017, pp. 1655--1661.

\bibitem{said1984testing}
S.~E. Said, D.~A. Dickey, Testing for unit roots in autoregressive-moving
  average models of unknown order, Biometrika 71~(3) (1984) 599--607.

\bibitem{williams1998urban}
B.~Williams, P.~Durvasula, D.~Brown, Urban freeway traffic flow prediction:
  application of seasonal autoregressive integrated moving average and
  exponential smoothing models, Transportation Research Record: Journal of the
  Transportation Research Board~(1644) (1998) 132--141.

\bibitem{hyndman2007automatic}
R.~J. Hyndman, Y.~Khandakar, Automatic time series for forecasting: The
  forecast package for R, Monash University, Department of Econometrics and
  Business Statistics, 2007.

\bibitem{hyndman2014forecasting}
G.~A. R.~J.~Hyndman, Forecasting: Principles and practice, Otexts.com.

\bibitem{zhang2011seasonal}
N.~Zhang, Y.~Zhang, H.~Lu, Seasonal autoregressive integrated moving average
  and support vector machine models: prediction of short-term traffic flow on
  freeways, Transportation Research Record 2215~(1) (2011) 85--92.

\bibitem{lippi2013short}
M.~Lippi, M.~Bertini, P.~Frasconi, Short-term traffic flow forecasting: An
  experimental comparison of time-series analysis and supervised learning, IEEE
  Transactions on Intelligent Transportation Systems 14~(2) (2013) 871--882.

\bibitem{habtemichael2016short}
F.~G. Habtemichael, M.~Cetin, Short-term traffic flow rate forecasting based on
  identifying similar traffic patterns, Transportation research Part C:
  emerging technologies 66 (2016) 61--78.

\bibitem{zhu2016short}
Z.~Zhu, B.~Peng, C.~Xiong, L.~Zhang, Short-term traffic flow prediction with
  linear conditional gaussian bayesian network, Journal of Advanced
  Transportation 50~(6) (2016) 1111--1123.

\bibitem{akagi2018fast}
Y.~Akagi, T.~Nishimura, T.~Kurashima, H.~Toda, A fast and accurate method for
  estimating people flow from spatiotemporal population data, in: IJCAI, 2018,
  pp. 3293--3300.

\bibitem{liu2018think}
Z.~Liu, P.~Zhou, Z.~Li, M.~Li, Think like a graph: Real-time traffic estimation
  at city-scale, IEEE Transactions on Mobile Computing PP~(99) (2018) 1--1.

\bibitem{lecun1998gradient}
Y.~LeCun, L.~Bottou, Y.~Bengio, P.~Haffner, Gradient-based learning applied to
  document recognition, Proceedings of the IEEE 86~(11) (1998) 2278--2324.

\bibitem{krizhevsky2012imagenet}
A.~Krizhevsky, I.~Sutskever, G.~E. Hinton, Imagenet classification with deep
  convolutional neural networks, in: Advances in Neural Information Processing
  Systems, 2012, pp. 1097--1105.

\bibitem{williams1989learning}
R.~J. Williams, D.~Zipser, A learning algorithm for continually running fully
  recurrent neural networks, Neural Computation 1~(2) (1989) 270--280.

\bibitem{sutskever2014sequence}
I.~Sutskever, O.~Vinyals, Q.~V. Le, Sequence to sequence learning with neural
  networks, in: Advances in Neural Information Processing Systems, 2014, pp.
  3104--3112.

\bibitem{jiang2013review}
S.~Jiang, G.~A. Fiore, Y.~Yang, J.~Ferreira~Jr, E.~Frazzoli, M.~C.
  Gonz{\'a}lez, A review of urban computing for mobile phone traces: Current
  methods, challenges and opportunities, in: Proceedings of the 2nd ACM SIGKDD
  International Workshop on Urban Computing, ACM, 2013, pp. 1--9.

\bibitem{calabrese2015urban}
F.~Calabrese, L.~Ferrari, V.~D. Blondel, Urban sensing using mobile phone
  network data: a survey of research, ACM Computing Surveys 47~(2) (2015)
  1--20.

\bibitem{li2015traffic}
Y.~Li, Y.~Zheng, H.~Zhang, L.~Chen, Traffic prediction in a bike-sharing
  system, in: Proceedings of the 23rd SIGSPATIAL International Conference on
  Advances in Geographic Information Systems, ACM, 2015, pp. 33:1--33:10.

\bibitem{he2016deep}
K.~He, X.~Zhang, S.~Ren, J.~Sun, Deep residual learning for image recognition,
  in: Proceedings of the IEEE International Conference on Computer Vision and
  Pattern Recognition, 2016, pp. 770--778.

\bibitem{deri2016big}
J.~A. Deri, F.~Franchetti, J.~M. Moura, Big data computation of taxi movement
  in new york city, in: 2016 IEEE International Conference on Big Data (Big
  Data), IEEE, 2016, pp. 2616--2625.

\bibitem{zhan2017citywide}
X.~Zhan, Y.~Zheng, X.~Yi, S.~V. Ukkusuri, Citywide traffic volume estimation
  using trajectory data, IEEE Transactions on Knowledge and Data Engineering
  29~(2) (2017) 272--285.

\bibitem{zhao2017lstm}
Z.~Zhao, W.~Chen, X.~Wu, P.~C. Chen, J.~Liu, Lstm network: a deep learning
  approach for short-term traffic forecast, IET Intelligent Transport Systems
  11~(2) (2017) 68--75.

\bibitem{duan2018improved}
Z.~Duan, Y.~Yang, K.~Zhang, Y.~Ni, S.~Bajgain, Improved deep hybrid networks
  for urban traffic flow prediction using trajectory data, IEEE Access 6 (2018)
  31820--31827.

\bibitem{jiang2019geospatial}
W.~Jiang, L.~Zhang, Geospatial data to images: A deep-learning framework for
  traffic forecasting, Tsinghua Science and Technology 24~(1) (2019) 52--64.

\bibitem{zhang2019flow}
J.~Zhang, Y.~Zheng, J.~Sun, D.~Qi, Flow prediction in spatio-temporal networks
  based on multitask deep learning, IEEE Transactions on Knowledge and Data
  Engineering (2019) 1--1.

\bibitem{liu2019deeppf}
Y.~Liu, Z.~Liu, R.~Jia, Deeppf: A deep learning based architecture for metro
  passenger flow prediction, Transportation Research Part C: Emerging
  Technologies 101 (2019) 18--34.

\bibitem{ma2018parallel}
X.~Ma, J.~Zhang, B.~Du, C.~Ding, L.~Sun, Parallel architecture of convolutional
  bi-directional lstm neural networks for network-wide metro ridership
  prediction, IEEE Transactions on Intelligent Transportation Systems (2018)
  1--11.

\bibitem{ning2018st}
Y.~Ning, Y.~Huang, J.~Li, Q.~Liu, D.~Yang, W.~Zheng, H.~Liu, St-drn: Deep
  residual networks for spatio-temporal metro stations crowd flows forecast,
  in: Proceedings of 2018 International Joint Conference on Neural Networks
  (IJCNN), IEEE, 2018, pp. 1--8.

\bibitem{nam2017model}
D.~Nam, H.~Kim, J.~Cho, R.~Jayakrishnan, A model based on deep learning for
  predicting travel mode choice, in: Proceedings of the Transportation Research
  Board 96th Annual Meeting Transportation Research Board, Washington, DC, USA,
  2017, pp. 8--12.

\bibitem{chai2018bike}
D.~Chai, L.~Wang, Q.~Yang, Bike flow prediction with multi-graph convolutional
  networks, in: Proceedings of the 26th ACM SIGSPATIAL International Conference
  on Advances in Geographic Information Systems, ACM, 2018, pp. 397--400.

\bibitem{walraven2016traffic}
E.~Walraven, M.~T. Spaan, B.~Bakker, Traffic flow optimization: A reinforcement
  learning approach, Engineering Applications of Artificial Intelligence 52
  (2016) 203--212.

\bibitem{jiang2018reinforcement}
Z.~Jiang, W.~Fan, W.~Liu, B.~Zhu, J.~Gu, Reinforcement learning approach for
  coordinated passenger inflow control of urban rail transit in peak hours,
  Transportation Research Part C: Emerging Technologies 88 (2018) 1--16.

\bibitem{wang2018crowd}
L.~Wang, X.~Geng, X.~Ma, F.~Liu, Q.~Yang, Crowd flow prediction by deep
  spatio-temporal transfer learning, arXiv preprint arXiv:1802.00386.

\bibitem{wang2018road}
B.~Wang, Z.~Yan, J.~Lu, G.~Zhang, T.~Li, Road traffic flow prediction using
  deep transfer learning, in: Data Science and Knowledge Engineering for
  Sensing Decision Support: Proceedings of the 13th International FLINS
  Conference (FLINS 2018), Vol.~11, World Scientific, 2018, pp. 331--338.

\end{thebibliography}

\end{document}